\documentclass{article} 

\usepackage{iclr2022_conference,times}

\usepackage{amsmath,amsfonts,bm}

\def\eqref#1{equation~\ref{#1}}

\def\1{\bm{1}}

\DeclareMathAlphabet{\mathsfit}{\encodingdefault}{\sfdefault}{m}{sl}
\SetMathAlphabet{\mathsfit}{bold}{\encodingdefault}{\sfdefault}{bx}{n}

\usepackage{hyperref}
\usepackage{url}

\usepackage{booktabs}
\usepackage{graphicx}
\usepackage{algorithm}
\usepackage{algpseudocode}
\usepackage{algorithmicx}

\usepackage{multicol}

\usepackage{multirow}
\usepackage{amssymb}
\usepackage{amsmath}
\usepackage{subcaption}
\usepackage{amsthm}
\newtheorem{mythm}{Theorem}
\usepackage{threeparttable}

\title{Enhanced countering adversarial attacks via input denoising and feature restoring}

\iclrfinalcopy
\author{Yanni Li \& Wenhui Zhang \& Jiawei Liu \& Xiaoli Kou \& Hui Li \& Jiangtao Cui \\
School of Computer Science and Technology\\
Xidian University\\
\texttt{yannili@mail.xidian.edu.cn,wenhui110920@gmail.com}\\
\texttt{liujw@stu.xidian.edu.cn,\{xlkou,hli,cuijt\}@xidian.edu.cn}
}

\begin{document}

\maketitle

\begin{abstract}
Despite the fact that deep neural networks (DNNs) have achieved prominent performance in various applications, it is well known that DNNs are vulnerable to adversarial examples/samples (AEs) with imperceptible perturbations in clean/original samples. To overcome the weakness of the existing defense methods against adversarial attacks, which damages the information on the original samples, leading to the decrease of the target classifier accuracy, this paper presents an enhanced countering  adversarial attack method IDFR (via \underline{I}nput \underline{D}enoising and \underline{F}eature \underline{R}estoring)
. The proposed IDFR is made up of an enhanced input denoiser (ID) and a hidden lossy feature restorer (FR) based on the convex hull optimization. Extensive experiments conducted on benchmark datasets show that the proposed IDFR outperforms the various state-of-the-art defense methods, and is highly effective for protecting target models  against various adversarial black-box or white-box attacks. \footnote{Souce code is released at: \href{https://github.com/ID-FR/IDFR}{https://github.com/ID-FR/IDFR}}
\end{abstract}

\vspace{-0.7em}
\section{INTRODUCTION}
In recent years, we have witnessed that deep neural networks (DNNs) have achieved prominent performance in various applications, such as  autonomous vehicles, robotics, network security, image/speech recognition, natural language processing, etc. However, a large number of applications and theoretical researches \citep{r21-1,r2,r1,r5,r2-1,r2-2}  show that the DNNs are vulnerable to adversarial examples/samples (AEs) with imperceptible perturbations in clean/original samples, namely  DNNs' vulnerability against adversarial attacks. \cite{r3} proposed the concept of AE, which means that by adding a slight perturbation to the input data, CNNs could easily mis-classify the AEs with high confidence, while human eyes cannot  distinguish these subtle differences. The CNNs' vulnerability raises both theory-wise issues about the interpret-ability of deep learning and  application-wise issues when deploying the CNNs in security-sensitive applications \citep{r33}. 

To overcome the above issues, many methods of defending the AEs have been proposed, which can  roughly fall into the following three categories. The first category is to enhance the robustness of CNNs itself. Adversarial training methods \citep{r6,r10-20,r9,r7,r8,r26,r28,r32,r29} are the representatives among them. These methods inject AEs into the training data to retrain the CNNs. Label smoothing methods, e.g., methods  \citep{r10-22,r36} convert  one-hot labels to soft targets, also belonging to this class. Another one refers to the various pre-processing/purification methods \citep{r13,r14,r21,r10,r27}, which focus on shifting the AEs back to their clean data representations, namely purification. It's worth noting that the self-supervised learning methods \citep{r33-3,r33-4,r15,r33-6,r33} have emerged recently and formed the third category. Studies have shown that the self-supervised learning can improve adversarial robustness more recently. These methods generally combine self-supervised learning with adversarial training to achieve the adversarial purification.

Although many adversarial defense  methods have achieved their  competitive robustness performance , we have observed the fact that all of the existing methods suffer from a key weakness, i.e., they focus on defending against adversarial attacks, but ignore the information loss/the partial knowledge forgetting learned by the CNN model for clean samples during the adversarial training/purification process, which leads to a decrease in the target classifier accuracy, which is a serious secondary disaster caused by the adversarial attacks to CNNs robustness. We suggest that an effective adversarial defense method must address this issue. To bridge the gap and to overcome the weakness,  this paper presents a novel and enhanced countering adversarial attack method IDFR (via \underline{I}nput \underline{D}enoising and \underline{F}eature \underline{R}estoring). The proposed IDFR consists of an input denoiser (ID) and a hidden lossy feature restorer (FR). The ID is used for enhanced denoising/pre-processing input AEs, and the FR is used for restoring the hidden lossy features of the clean samples based on the convex hull optimization.

Compared with the above three types of the state-of-the-art  
representative adversarial defense methods, e.g., the methods \citep{r11,r14,r12,r10,r15,r26,r33}, etc., our proposed IDFR has the following distinct advantages. First, based on a new designed ID with a U-Net convolutional network \citep{r19}, and with the joint training of AEs and enhanced clean samples, i.e., clean samples with Gaussian disturbance augmentation, our IDFR achieves stronger input denoising capability, and its ID can effectively prevent over-fitting to AEs. Second, with the convex hull optimization \citep{r48}, the linear convex combinations of the hidden features of the denoising AEs and clean samples with misclassification are devised to train the FR of the IDFR, leading  to effectively recovering the lossy information on the clean samples and avoiding the decrease of the target classifier accuracy. Third, both the components ID and FR in the IDFR are pluggable, and they can be transferred across different target models. Extensive experiments conducted on benchmark datasets show that the performance of our proposed method IDFR greatly outperforms that of the state-of-the-art defense methods, and is consistently effective in protecting target classifiers against AEs and lossy clean samples.


\section{Preliminary and related work}
\vspace{-0.5em}
\subsection{Preliminary}
\textbf{Adversarial examples AEs} \citet{r38} found adversarial attack phenomenon, and \citet{r3} proposed AEs to fool DNNs. Adding a subtle perturbation to the input of a DNN will produce an error output with high confidence, while human eyes cannot recognize the difference. Suppose that there are a target model $f_{\theta}$ and an original/clean example $x$, which can be correctly classified by the model, i.e., $f_{\theta}(x)=y$, where $y$ is the true class label of $x$. However, it is possible to construct an AE $x'$ which is perceptually indistinguishable from $x$ but is classified incorrectly, i.e., $f_{\theta}(x') \ne y$. 

\textbf{Problem statement} Consider an encoder $f_{enc}$:  $h_x=f_{enc}(x;\theta_{enc})$, a classifier $f_{cls}(h_x;\theta_{cls})$ on top of the representation/embedding $h_x$, and the target model $f= f_{cls} \circ f_{enc}$, a composition of the encoder and the classifier. The adversarial denoising/purification problem can be formulated as follows: for an adversarial example $x'$ and its clean counterpart $x$, our denoising/purification strategies $\pi_1$ and $\pi_2$ aim to: 1) find $x^{*}=\pi_1(x')$ that is as close to the clean/original example $x$ as possible: $x^{*} \rightarrow x$, and 2) achieve $y_{h_{x}}= \pi_{2}(y_{h_{\tilde{x}}}$), i.e.,  $y_{h_{\tilde{x}}} \rightarrow y_{h_{x}} (=y)$, where $\tilde{x}$ is the lossy example corresponding to a  clean example $x$ partially damaged by the 
previous denoising operation, $h_{\tilde{x}}=f_{enc}(\tilde{x})$, and $y_{h_{\tilde{x}}}=f_{cls}(h_{\tilde{x}})$. However, this problem is under-determined as different clean examples can share the same adversarial counterpart, i.e., there might be multiple or even infinite solutions for $x$. Thus, we consider the following relaxation 
\begin{equation}\label{eq:problem}
\begin{aligned}
&\min_{\pi_1} \mathcal{L}_{cls} \bigl( f_{cls}(x^*), y \bigr) + \min_{\pi_2} \mathcal{L}_{cls} \bigl( f_{cls}(h_{\tilde{x}}), y \bigr) \\
&\text{s.t.}\quad
\Vert x^* - x' \Vert \leq \epsilon, \;\; x^* = \pi_1(x'), \;\; y_{h_{x}}= \pi_{2}(y_{h_{\tilde{x}}})=y
\end{aligned}
\end{equation}
where $\mathcal{L}_{cls}$ is the cross entropy loss for the  classifier and $\epsilon$ is the budget of adversarial perturbation.

It is worth noting that, to our knowledge, this is the first time we presented the complete formal definition of the problem. In addition, the existing adversarial defense methods only achieve the above adversarial defense strategy $\pi_1$ well, but ignore/do not achieve the $\pi_2$.

\begin{figure}[h]
    \centering
    \includegraphics[scale=0.85]{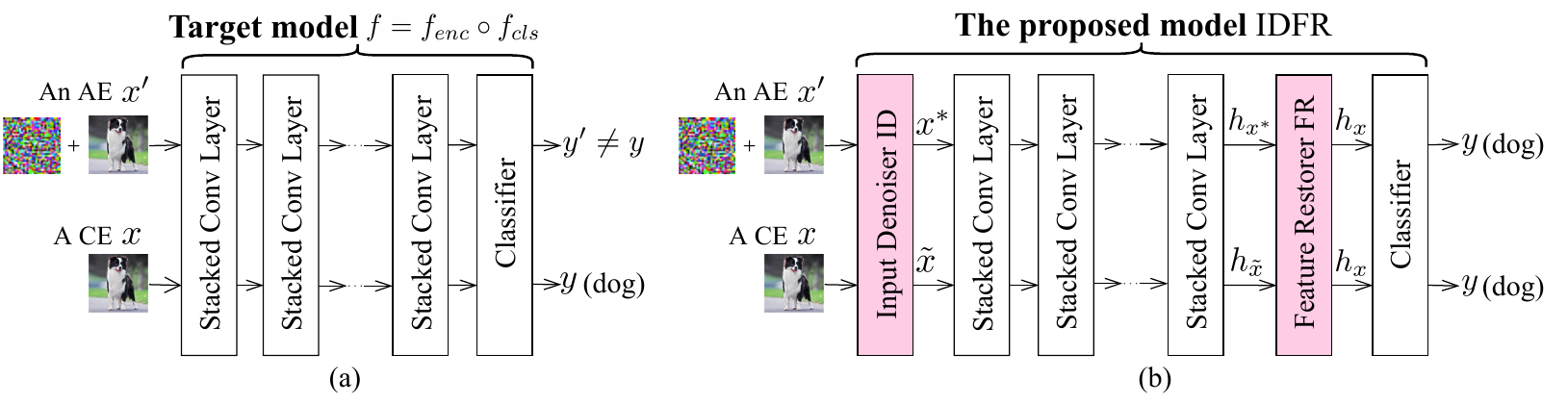}
    \caption{The schematic diagram of the target model (a) and the proposed model IDFR (b).}
    \label{fig1}
\end{figure}

\vspace{-0.7em}
\subsection{Related work}\label{subsec:related-work}
In recent years, for the severe challenge of adversarial attacks to CNNs, many adversarial defense methods have been proposed, which can be roughly divided into the following three categories. 
As space does not allow for a comprehensive literature study, we refer readers to \citet{r5,r1-3,r1-1,r31,r1-2} for a survey of these works. Hereby, we only focus on some latest state-of-the-art methods relevant to our study.

\textbf{Adversarial training} Adversarial training aims to improve CNNs' robustness through data augmentation, where the target model  is trained on adversarial perturbed examples, i.e., AEs, instead of clean/original training samples \citep{r6,r9,r7,r28,r32,r29}. By solving a min-max problem, the model learns a smoother data manifold and decision boundary, which improves robustness of the DNNs. However, the computational costs of the general adversarial training methods are high as strong adversarial examples are typically found in an iterative manner with heavy gradient calculation. To overcome the weakness, \citet{r26} revealed that adversarial training with the fast gradient sign method FGSM \citep{r22}, when combined with random initialization, is as effective as PGD-based training \citep{r34} but has a significantly lower cost.

\textbf{Adversarial denoising/purification} This kind of robust learning focuses on shifting the AEs back to the clean counterparts, namely adversarial denoising/purification. Representative methods are DCN \citep{r13}, HGD \citep{r11}, Defense-GAN \citep{r42}, ComDefend \citep{r10}, NPR \citep{r15}, Feature Squeezing \citep{r12}, JPEG Compression \& TVM \citep{r14}, ADP \citep{r27}, etc. \citet{r13} exploited a general DAE \citep{r11-17} to remove adversarial noises. \citet{r42} trained a GAN on clean examples and projected the AEs to the manifold of the generator, but the GAN was hard and inefficient to train. \citet{r10} proposed an end-to-end image compression model ComDefend to defend against AEs, and defeated the state-of-the-art defense models including the winner \citep{r11} of NIPS 2017 adversarial challenge. \citet{r14} introduced two defense models, i.e., JPEG Compression \& TVM, with their best defense eliminating 60\% of strong gray-box and 90\% of strong black-box attacks by their defense methods. \citet{r27} proposed a novel adversarial purification method ADP based on an energy-based model trained with denoising score-matching, and the proposed ADP could  quickly purify attacked images within a few steps. 

\textbf{Self-supervised learning} Self-supervised learning aims to learn intermediate representations of unlabeled data that are useful for unknown downstream tasks. More recently, studies have shown how self-supervised learning can improve adversarial robustness, leading to the new type of self-supervised learning methods. \citet{r33-3} discovered that adversarial attacks fool the networks by shifting latent representation to a false class. \citet{r33-4} observed that PGD adversarial training along with an auxiliary rotation prediction task improved robustness. \citet{r15} utilized feature distortion as a self-supervised signal to find transferable attacks that are generalized across different architectures and tasks. These methods typically combine self-supervised learning with adversarial training, and thus the computational cost is still high. In contrast, with a variety of self-supervise signals as auxiliary objectives, SOAP \citep{r33} achieved a competitive robust accuracy by test-time purification, but its test-time computation and accuracy still leaved rooms for improvement.

Although the above various adversarial defense methods have achieved a competitive robust accuracy, we observe the fact that these methods suffer from a serious  weakness, i.e., they focus mainly on defending against AEs attacks, but ignore the information loss/the partial knowledge forgetting learned by the CNNs model of the clean samples during the adversarial training/purification process, which leads to a decrease in the target  classifier accuracy. In this paper, we aim to bridge the gap and to overcome the key weakness. Specifically, we try to obtain the optimal solution of the problem   against adversarial attacks (see Subsec \ref{subsec:related-work} problem statement). 

\section{Input denoising and feature restoring method: IDFR}

\vspace{-0.5em}
\subsection{Overview of the proposed IDFR}

To eliminate the key weakness of existing methods, we propose a novel adversarial defense method IDFR, which consists of two independent but collaborative models, i.e., the input denoiser (ID) and the hidden lossy feature restorer (FR). The structure of the IDFR is shown in Fig. \ref{fig1}.

More specifically, the ID can transform an adversarial image to its clean version, which is a pre-processing module and does not modify the target model structure during the whole process, while the FR is used to restore those partially damaged clean samples \{$\tilde{x}$\} so that they can be correctly classified under the constrain of the invariance of the denoised samples \{$x^{*}$\} produced by the ID. For efficiently  recovering these \{$\tilde{x}$\}, the input of the FR takes the embedding of the $\tilde{x}$ generated by the target model encoder $f_{enc}$ rather than $x$ itself. The reason is that the dimension of the embedding is much smaller than that of $x$. Therefore,  our FR module is designed as a pluggable component after the output of the target model encoder  $f_{enc}$. Ultimately, both the ID and FR of our IDFR jointly enhance/protect the classifier's robustness of the target model. The design ideas and the theoretical basis behind our proposed IDFR will be detailed in following subsections.   
    
\vspace{-0.7em}
\subsection{Input denoiser ID}
\textbf{Network architecture of the ID} Although DAE (\underline{D}enoising \underline{A}uto\underline{E}ncoder) \citep{r10-20} is a popular denoising model, it has a bottleneck structure between the encoder and decoder. This bottleneck may constrain the transmission of fine-scale information necessary for reconstructing high-resolution images. We introduce the U-Net convolutional network \citep{r19} as the network model of our ID. It is due to the fact that the U-Net  network can overcome the bottleneck of the DAE. the U-Net  network has both a contracting path to extract contextual information and a symmetric expanding path to capture precise local information. Our ID network architecture is shown in Appendix \ref{appendix:IDFRstructure}. 

\textbf{Training of the ID} Let the ID model be $I_w(\cdot)$ with the input  an AE $x'$ and/or a clean example (CE) $x$, where $w$ is the parameters of the ID. We expect that the output of the ID is the denoised sample $x^{*}$ of $x$. Thus, the optimal objective of the ID is designed as follows 
\begin{equation}\label{ID-indicate}
    \mathcal{L}_\text{ID}=\arg\max_w \sum  \mathbb{I} \Bigl( f_{cls} \bigl( I_w(x, x') \bigr) = y\Bigr)
\end{equation}
where $f_{cls}$ is the classifier of the target model $f$, $y$ is the ground truth class label of both $x$ and $x'$, and  $\mathbb{I}(\cdot)$ is an indicating function.  $\mathbb{I}(true)=1$, otherwise  $\mathbb{I}(false)=0$. As the $\mathbb{I}(\cdot)$ would result in  the issue of gradient disappearances with the back propagation optimization, we replace it with the cross entropy loss, leading to the following Eq. \ref{ID-indicate}

\begin{equation}\label{eq:ID-cross-entropy}
    \mathcal{L}_\text{ID}=\arg\min_w \Bigl(- \sum y\log f_{cls} \bigl( I_w(x, x') \bigr) = y \Bigr)
\end{equation}

It's worth noting that our ID adopts the jointly training with both AEs and clean examples to avoid its over-fitting the AEs. Moreover, when the adversarial disturbance of the AE is large, i.e., the adversarial attack is strong, the ID's learning will favor this strong AE, i.e., the ID would over-fitting the strong AE. To address the issue, the  multi-round Gaussian perturbation data enhancement method \citep{r18} is employed to obtain the enhanced counterpart $x_e$ of $x$, which greatly improves the ID model's learning to clean samples $x$. Therefore, considering the above two factors, the ID model adopts the joint training with such training datasets including both \{$x'$\} and \{$x_e$\}. By being equipped with the U-Net network and the enhanced jointly  training strategy, our ID achieves a stronger ability of the adversarial denoising and model generalization than that of the existing denoising/purification strategies. The ID training algorithm is shown in Algorithm \ref{algo:ID}.

\vspace{-0.4em}
\subsection{Lossy features restorer FR}
\vspace{-0.4em}
\subsubsection{The design motivation of the FR}

As the above discussion, we have known that there may be residual noises/information damaged in the generated outputs $x^{*}$/$x$ (denoted as $\tilde{x}$) when $x$ and $x'$ are fed into the ID, and that the effect of the residual noises/information damaged in the  $x^{*}$/$\tilde{x}$ increase rapidly along with the depth of the target model. Note that this is the infrasonic disaster caused by adversarial attacks, which also poses a great threat to the robustness of the target model. Therefore, it is essential that such disaster/threat should be dealt with effectively. To this end, the network structure and learning/training process of the FR should be well-designed. Similar to the ID network, the FR network also utilizes the U-Net network, whose structure and hyperparameters are shown in Appendix \ref{appendix:IDFRstructure}.

\subsubsection{Training of the FR}  By the theoretical study and extensive experiments, we have found that some of these outputs \{$x^{*}$\} and \{$\tilde{x}$\} of the ID can be correctly classified by the target model's classifier $f_{cls}$, while a few others get mis-classified. As we pointed out earlier, this phenomenon is an infrasonic disaster caused by adversarial attacks that must be contained and resolved well. From this point of view, we devised a novel learning/ training strategy of the FR. For the efficiency of our FR learning/training, we only consider the embeddings $h_{x^{*}}$ and $h_x$/$h_{\tilde{x}}$ generated by the target model encoder, i.e., $h_{x^{*}}=f_{enc}(x^{*}), h_x=f_{enc}(x)$, and $h_{\tilde{x}}=f_{enc}(\tilde{x})$ as the input of our FR, which is due to the fact that the dimension of the embedding is much smaller than that of a raw data itself. For clarifying our proposed FR's learning/training algorithm, following definitions and Theorem \ref{theorem1} are first introduced.

\textbf{Definition 1: Classified space of data.} For a specific category of input data, we define $p_1$ as the space where the embeddings, i.e., $h_{x^{*}}$ and $h_x$/$h_{\tilde{x}}$, can be correctly classified by the target classifier $f_{cls}$, and $p_2$ as the space where embeddings are misclassified. The complete classification space of the embeddings is defined as $p$, i.e., $p=p_1 \cup p_2$. To simplify the notation, those embeddings falling into $p_1$ are uniformly denoted as $\hat{\textbf{x}}=\{h_x\}$, while the others falling into $p_2$ are represented by $\hat{\textbf{x}}'=\{h_{\tilde{x}}\}$. The schematic diagram of the classified spaces and the convex hull is shown Fig. \ref{fig:convex}.

\textbf{Definition 2: Convex hull $Co(\cdot)$.} For a given data set $\textbf{x} \in \mathbb{R}$, the intersection of all convex sets containing the $\textbf{x}$ is called the convex hull of $\textbf{x}$, denoted as $Co(\textbf{x})$, which can be constructed from a convex combination of all points in $\textbf{x}$ \citep{r48}.

 Based on the theory of convex hull optimization \citep{r48}, the following Theorem \ref{theorem1} can be introduced and proved (see Appendix \ref{appendix:proof}).

\begin{mythm}\label{theorem1}
A convex combination of any two points in a $Co(\cdot)$ remains in the $Co(\cdot)$ region.
\end{mythm}

According to the above definitions and Theorem \ref{theorem1}, we can obtain  the convex hull $Co(\hat{\textbf{x}} \cup \hat{\textbf{x}}')$ of both  $\hat{\textbf{x}}=\{h_x\}$ and  $\hat{\textbf{x}}'=\{h_{\tilde{x}}\}$ as follows
\begin{equation}\label{eq:convex-def}
    Co(\hat{\textbf{x}} \cup \hat{\textbf{x}}')=\{\alpha h_x + (1-\alpha)h_{\tilde{x}} | h_x \in \hat{\textbf{x}}, h_{\tilde{x}} \in \hat{\textbf{x}}'\}
\end{equation}
where $\alpha$ is an equilibrium coefficient between 0 and 1. That is, the smaller the value of $\alpha$, the closer to $h_{\tilde{x}}$ the convex hull $Co(\hat{\textbf{x}} \cup \hat{\textbf{x}}')$ is, and vice versa.

How can we use the FR to restore the clean sample with damaged information and the AE with residual noise, i.e., $\hat{\textbf{x}}'=\{h_{\tilde{x}}\}$, so that they can finally be correctly classified? This question  is essentially equivalent to this: How can we pull the $\hat{\textbf{x}}'= \{ h_{\tilde{x}} \}$ from the misclassified region $p_2$ back into the correctly classified region $p_1$ by the FR? To this end, based on the convex hull optimization \citep{r48}, and inspired by the method mixup \citep{r20}, i.e., a data augmentation method, we devised the following optimal learning/training strategy for our FR model, denoted as $F_{\pi}(\cdot)$, where $\pi$ represents the  parameters of the FR model. Specifically, 1) to pull the $\hat{\textbf{x}}'= \{ h_{\tilde{x}} \}$ from the misclassified region $p_2$ back into the correctly classified region $p_1$, and 2) to maintain those correctly classified data $\hat{\textbf{x}}= \{h_x\}$ still in its classified area $p_1$, we proposed the cross entropy loss of the FR as follows. 
\begin{equation}\label{eq:FRloss}
    \mathcal{L}_{FR}=\arg\min_\pi \bigl( -\sum y\log f_{cls} \bigl( F_\pi(h_{\tilde{x}}) \bigr)-\sum y\log f_{cls} \bigl( F_\pi(h_x) \bigr) \bigr)
\end{equation}
where the first part in  Eq. \ref{eq:FRloss} is used to accomplish the first objective above, and the second part is used to achieve the second objective. It is worth mentioning that in order to achieve its end and to improve the generalization ability of the FR model, the Eq. \ref{eq:FRloss}  is not directly used in the training process of our FR model, but another form Eq. \ref{eq:convex} is utilized.

Since the data $\hat{\textbf{x}}=\{h_x\}$ and/or $\hat{\textbf{x}}'=\{h_{\tilde{x}}\}$ in the regions $p_1$ and $p_2$, respectively, are all in the $Co(\hat{\textbf{x}} \cup \hat{\textbf{x}}')$, based on the convex hull optimization theory \citep{r48}, the Eq. \ref{eq:FRloss} can be converted into Eq. \ref{eq:convex}.
\begin{equation}\label{eq:convex}
    \mathcal{L}_{FR}=\arg\min_\pi \bigl( -\sum y\log f_{cls} \bigl( F_\pi(\alpha h_x + (1-\alpha)h_{\tilde{x}} \bigr) \bigr) , h_{x}, h_{\tilde{x}} \in Co(\hat{\textbf{x}} \cup \hat{\textbf{x}}'), \alpha \in [0,1]
\end{equation}
Equipped with the loss function $\mathcal{L}_{FR}$ and the convex hull $Co(\hat{\textbf{x}} \cup \hat{\textbf{x}}')$, the FR can achieve its end well. The training process of the proposed FR is shown in Algorithm \ref{algo:FR}.

From the above discussion, it can be clearly seen that, different from all the existing adversarial defense methods, with the ID and FR, our proposed method IDFR solves the problem (see Eq. \ref{eq:problem}) of the defense against adversarial attacks well, which is also fully verified by our extensive experimental results. 

\begin{minipage}{\dimexpr.5\textwidth-.5\columnsep}
\begin{algorithm}[H]
  \caption{The training process of ID}
  \label{algo:ID}
  \begin{algorithmic}[1]
      \Require Target model $f_\theta(\cdot)$, produce the $\epsilon$ model $A(\cdot)$, ID model $I_w(\cdot)$, training epochs $T$, training datasets $\{x\}$ and $\textbf{x}_e$, learning rate $\lambda$
      \Ensure ID model $I_w$
      \For{each $x$ in \{$x$\}} 
          \State $x'\leftarrow x+\epsilon, y'\leftarrow y$
          \State $\textbf{x}'\leftarrow \textbf{x}' \cup \{ (x',y') \}$
      \EndFor
      \For{each $i \in 1, \cdots, T$}
          \For{each $(x',y)$ in $\textbf{x}'\cup \textbf{x}_e$}
              \State Update $w$ by Eq. \ref{eq:ID-cross-entropy} with Adam
          \EndFor
      \EndFor
      \State \Return $I_w$
  \end{algorithmic}
\end{algorithm}
\end{minipage}\hfill%
\begin{minipage}{\dimexpr.5\textwidth-.5\columnsep}
\begin{algorithm}[H]
  \caption{The training process of FR}
  \label{algo:FR}
  \begin{algorithmic}[1]
      \Require Target model $f_\theta(\cdot)$, pretrained ID model $I_w(\cdot)$, FR model $F_\pi(\cdot)$, Training epochs $T$, enhenced samples $\textbf{x}_e$, learning rate $\eta$
      \Ensure FR model $F_\pi(\cdot)$
      \For{each $i=1$ to $T$ }
          \For{$h_{x}, h_{\tilde{x}}$ in  $Co(\hat{\textbf{x}} \cup \hat{\textbf{x}}')$ }
                \State $\alpha \leftarrow \text{rand}(0,1)$
                \State Calculate $\alpha h_x$
                \State Calculate $(1-\alpha)h_{\tilde{x}}$
                \State Calculate $Co(\cdot) =\alpha h_x + (1-\alpha)h_{\tilde{x}}$
                \State Update $\pi$ by Eq. \ref{eq:convex} with Adam 
          \EndFor
      \EndFor
      \State \Return $F_\pi$
  \end{algorithmic}
\end{algorithm}
\end{minipage}

\section{Experiments}

In this study, all experiments were conducted on the server: Intel Xeon(R) Gold 5115 CPU @ 2.40GHz, 97GB RAM and NVIDIA Tesla P40 graphics processor.

\subsection{Target models, baselines, datasets and experimental settings}

\textbf{Target models and baselines} 
To fully evaluate the performance of our proposed IDFR, in this paper, the following various typical CNN models, i.e., VGG16 \citep{r49}, ResNet18 \citep{r50}, InceptionV3 \citep{r51} and ResNet50 \citep{r50}, are employed as the target models, seven methods belong to the three different types of the state-of-the-art adversarial defense methods, i.e., ComDefend \citep{r10}, NPR \citep{r15}, Feature Squeezing \citep{r12}, JPEG Compression \& TVM \citep{r14}, ADP \citep{r27}, Fast AT\citep{r22} and SOAP \citep{r33} are used as experimental baselines, where the ComDefend, Feature Squeezing, JPEG Compression \& TVM and ADP belong to the category of the adversarial denoising/purification, both NRP and SOAP improve adversarial robustness of the target models by self-supervised learning, while the  method Fast AT falls into the class of the adversarial training.

\textbf{Datasets} 1) MNIST \citep{r53}: this dataset consists of 70,000 28$\times$28 black-and-white images of handwritten digits from 0 to 9. We use 60,000/3000/7000 images for training/validation/testing respectively. 2) CIFAR-10 \citep{r54}: this dataset consists of 60,000 32$\times$32 color images of 10 classes, with 6000 images per class. We use 50,000/3000/7000 images for training/validation/testing respectively. 3) CIFAR-100 \citep{r54}: this dataset consists of 60,000 32$\times$32 color images of 100 classes, with 600 images per class. We use 50,000/3000/7000 images for training/validation/testing respectively. 4) ImageNet   \citep{r56}: this dataset consists of 30,000  224$\times$224 color images of 1000 classes, with 30 images per class. We use 25,000/5000/10,000 images for training/validation/testing respectively. 5) SVNH \citep{r55}: this dataset consists of 624,420  32$\times$32 color images, where 73,257/531,131/26,032 samples are used for training/extra training/testing, respectively. The extra training samples mean they are somewhat less difficult samples to use as extra training data.  

\textbf{Hyperparameter Settings}: The training epochs for our ID and FR models are set 100 and 80, respectively, and the learning rates of the models ID and FR are uniformly set to  0.01 for updating the models' parameters by the adam optimizer. For all the baselines' hyperparameter settings, we strictly follow the settings of the original papers.

\subsection{Experimental results}
\vspace{-0.4em}
\subsubsection{Attacking models}
The model used to generate adversarial attacks is called the attacking model, which can be a single model or an ensemble of models \citep{r10-20}. When the attacking model is the target model itself or contains the target model, the resulting attacks are white-box otherwise  black-box. An intriguing property of adversarial examples is that they can be transferred across different models \citep{r3,r22}. This property enables black-box attacks. Practical black-box attacks have been demonstrated in some real-world scenarios \citep{r23,r11-21,r2,r1,r2-1,r2-2,r2-3}, etc. As while-box attacks are less likely to happen in practical systems, defenses against black-box attacks are more desirable. Therefore, our  experiments focus on black-box attacks to fully evaluate the performance of our method IDFR and baselines.

Many adversarial attack models, e.g., \citep{r3,r22,r11-22,r4,r23} , are mainly used to generate various adversarial examples (AEs) for evaluating  the adversarial attacks performance of an adversarial defense method. \citet{r22} suggested that AEs can be caused by the cumulative effects of high dimensional model weights. Thus, they proposed a simple but widely used the adversarial attack algorithm, called FGSM (\underline{F}ast \underline{G}radient \underline{S}ign \underline{M}ethod), which only computes the gradients for once, and thus was much more efficient than  L-BFGS \citep{r3}. The attack model CW \citep{r4} demonstrates  that defensive distillation does not significantly increase the robustness of CNNs by introducing a new attack algorithm with $l_1, l_2$, and $l_\infty$, respectively, that are successful on both distilled and undistilled CNNs with 100\% probability. BIM  \citep{r47} explores methods of producing AEs on deep generative models such as the variational autoencoder (VAE) and the VAE-GAN, which can give 
three classes of attacks on the VAE and VAE-GAN architectures
and demonstrate them against networks trained on datasets MNIST,  SVHN and CelebA.

\vspace{-0.5em}
\subsubsection{Experimental results}
\textbf{Black-box attacks} In the  experiments, the adversarial attack models FGSM \citep{r22}, CW \citep{r4} and BIM \citep{r47}  are used to produce various AEs\footnote{where the models both CW and BIM are running 20 iterations with a step size of 0.03.}, which are abbreviated as FGSM, CW and BIM on the above benchmark datasets respectively, The $\epsilon$ (=4, 8 and 16, respectively, denoted as =4/8/16) represents the upper bound of the disturbance $\delta$ around $l_{\infty}$ norm, i.e., ${\Vert \delta \Vert}_{\infty} \leq \epsilon$. The experimental results under various target models and datasets are shown in Table \ref{tab:black-cifar10} to Table \ref{tab:imagenet}, respectively, and the statistical average values of the experiments are shown in Fig. \ref{tab:imagenet}. It is worth mentioning that as some methods, for the large-scale  dataset ImageNet, do not have the ability to process the large-scale images, or their training and/or testing are too inefficient to get experimental results, Table \ref{tab:imagenet} only shows the experimental results of the performance of our proposed IDFR and some comparable baselines on the ImageNet dataset. 

\begin{table}[h]
    \centering
    \caption{The performance results on  CIFAR-10 with black-box attacks}
    \resizebox{.98\textwidth}{20mm}{
     \begin{threeparttable}
        \begin{tabular}{lcccccccc}
        \toprule
          \multicolumn{1}{c}{Defense}  & \multicolumn{2}{c}{Clean Samples} & \multicolumn{2}{c}{FGSM($\epsilon=4/8/16$)} & \multicolumn{2}{c}{CW($\epsilon=4/8/16$)} &
          \multicolumn{2}{c}{BIM($\epsilon=4/8/16$)}\\ 
          \multicolumn{1}{c}{Method} & VGG16 & ResNet18 & VGG16 & ResNet18 & VGG16 & ResNet18 & VGG16 & ResNet18 \\ \midrule
          No Defense & 92.6 & 93.2 & 80.3/64.6/46.1 & 82.8/67.6/45.2 & 83.5/77.0/69.3 & 84.5/76.2/70.3 & 82.7/57.1/30.9 & 83.2/55.1/31.5  \\ 
          NRP & 79.4 & 78.7 & 76.7/74.9/72.2 & 74.8/72.5/68.1 & 78.5/78.5/77.5 & 77.1/76.7/75.8 & 77.4/74.8/71.4 & 75.6/72.5/68.3 \\ 
          Fast AT & 81.5 & 83.8 & 88.1/78.7/56.0 & 89.0/79.1/55.0 & 91.8/90.8/87.8 & 91.2/90.7/87.4 & 90.1/90.8/87.1 & 91.0/90.7/87.5 \\ 
          ComDefend & 85.2 & 87.1 & 82.7/78.8/70.8 & 82.2 77.3/69.9 & 84.5/84.5/82.6 & 84.6/84.5/81.9 & 83.2/80.3/76.6 & 84.0/79.8/74.9\\ 
          Feature Squeezing & 91.0 & 92.0 & 82.5/65.3/48.8 & 83.4/66.7/49.1 & 89.5/88.1/83.3 & 89.7/88.9/82.7 & 85.2/62.2/41.6 & 86.3/66.9/43.9 \\ 
          JPEG(q=25) & 71.0 & 74.1 & 66.6/63.0/54.3 & 66.2/64.1/55.0 & 70.1/69.8/68.2 & 69.9/69.9/68.5 & 67.5/64.2/60.2 & 68.2/63.1/59.4 \\ 
          JPEG(q=50) & 80.2 & 82.3 & 74.4/68.2/51.0 & 73.5/69.2/51.8 & 79.0/77.8/75.1 & 79.2/77.2/74.4 & 75.7/70.5/63.3 & 76.3/71.2/64.3 \\ 
          JPEG(q=75) & 85.9 & 84.9 & 78.4/68.5/47.5 & 79.2/69.7/48.1 & 84.4/83.0/78.5 & 85.9/84.2/78.8 & 80.5/72.9/58.8 & 81.0/73.4/58.9 \\ 
          TVM & 88.8 & 86.9 & 81.5/76.7/52.7 & 80.5/75.2/51.8 & 87.0/86.2/81.8 & 87.8/86.6/81.0 & 83.1/74.0/60.3 & 82.2/73.8/59.7 \\ 
          ADP & 86.4 & 83.6 & 77.6/74.1/70.4 & 75.1/71.6/66.2 & 78.7/79.5/78.8 & 77.8/77.1/76.2 & 77.8/74.7/70.1 & 75.5/72.4/68.7 \\ 
          SOAP & 89.7 & 90.1 & 71.9/71.8/71.9 & 70.7/70.6/70.1 & 71.8/71.0/70.2 & 70.3/69.9/69.3 & 71.8/71.9/71.2 & 71.6/71.9/71.8 \\ 
          IDFR(Ours) & \textbf{91.2} & \textbf{92.5} & \textbf{89.7/86.8/73.8} & \textbf{90.0/86.2/75.4} & \textbf{91.9}/ \textbf{90.9/90.6} & \textbf{91.7/90.8/90.2} & \textbf{90.4/91.7/88.6} & \textbf{91.2/90.9/89.1} \\ \bottomrule 
        \end{tabular}
      \begin{tablenotes}
        \footnotesize
           \item[1] p--the compression ratio parameter of the model JPEG. 
          \end{tablenotes} 
        \end{threeparttable}
    }
    \label{tab:black-cifar10}
\end{table}
\begin{table}[h]
    \centering
    \caption{The performance results on CIFAR-100 with black-box attacks}
    \resizebox{.98\textwidth}{20mm}{
    \begin{tabular}{lcccccccc}
    \toprule
          \multicolumn{1}{c}{Defense}  & \multicolumn{2}{c}{Clean Samples} & \multicolumn{2}{c}{FGSM($\epsilon=4/8/16$)} & \multicolumn{2}{c}{CW($\epsilon=4/8/16$)} &
          \multicolumn{2}{c}{BIM($\epsilon=4/8/16$)}\\ 
          \multicolumn{1}{c}{Method} & VGG16 & ResNet18 & VGG16 & ResNet18 & VGG16 & ResNet18 & VGG16 & ResNet18 \\ \midrule
      No Defense & 72.6 & 76.4 & 52.1/45.3/39.6 & 51.3/42.4/37.9 & 63.2/61.4/54.8 & 62.0/59.3/51.3 & 53.4/50.1/29.3 & 52.6/48.4/28.2  \\ 
      NRP & 65.2 & 65.4 & 65.4/62.1/61.2 & 69.2/68.5/68.3 & 68.1/68.7/66.3 & 71.8/70.9/70.3 & 69.3/67.7/59.3 & 72.1/70.7/66.4 \\ 
      Fast AT & 65.8 & 70.1 & 70.2/62.3/55.1 & 74.1/70.2/65.2 & 72.1/70.9/69.2 & 74.2/74.1/73.6 & 72.0/70.3/68.2 & 75.5/73.7/72.9 \\ 
      ComDefend & 68.7 & 71.4 & 66.2/61.4/60.2 & 70.2/69.8/67.2 & 70.0/69.7/65.2 & 73.8/71.2/69.1 & 68.2/66.3/60.1 & 73.2/71.7/67.5\\ 
      Feature Squeezing & 71.3 & 74.1 & 67.2/57.1/54.2 & 54.7/44.5/40.1 & 70.2/70.1/70.5 & 70.4/67.2/65.6 & 68.3/62.5/57.3 & 71.3/69.5/64.1 \\ 
      JPEG(q=25) & 55.4 & 55.7 & 47.2/42.3/39.1 & 49.2/41.3/39.1 & 57.2/57.1/56.8 & 56.1/54.1/50.4 & 57.7/55.4/50.0 & 58.2/56.2/51.4 \\ 
      JPEG(q=50) & 58.8 & 60.5 & 49.2/43.7/38.2 & 51.5/44.7/42.7 & 64.2/63.4/62.7 & 65.2/62.6/62.1 & 66.0/61.6/58.2 & 65.1/62.4/59.7 \\ 
      JPEG(q=75) & 68.2 & 71.8 & 58.9/47.2/45.2 & 55.2/47.5/45.4 & 69.3/68.9/67.2 & 68.3/67.3/63.5 & 67.3/58.2/45.5 & 63.3/59.2/47.5 \\ 
      TVM & 69.1 & 72.3 & 61.7/51.2/49.7 & 57.1/49.3/51.6 & 70.1/68.2/67.7 & 69.8/68.2/63.1 & 67.9/61.3/50.2 & 64.2/60.8/51.7 \\ 
      ADP & 67.9 & 69.2 & 62.7/62.6/62.4 & 64.4/64.2/63.8 & 68.7/68.9/67.3 & 70.4/70.5/69.4 & 64.7/63.5/62.3 & 66.7/66.8/65.1 \\ 
      SOAP & 62.1 & 65.6 & 59.7/58.5/57.7 & 62.6/61.2/60.9 & 65.8/65.3/64.7 & 67.6/67.5/66.6 & 62.7/62.8/61.3 & 62.2/61.6/61.1 \\ 
      IDFR(Ours) & \textbf{71.8} & \textbf{74.8} & \textbf{71.2/68.3/63.7} & \textbf{75.5/72.3/68.7} & \textbf{72.2/71.2/70.6} & \textbf{74.9/74.7/74.2} & \textbf{73.4/71.8/69.5} & \textbf{75.5/74.1/73.3} \\ \bottomrule
    \end{tabular}
    }
    \label{tab:cifar100}
\end{table}

\begin{figure}[!htb]
    \centering
    \begin{minipage}{0.45\textwidth}
        \centering
         \includegraphics[width=0.77\textwidth]{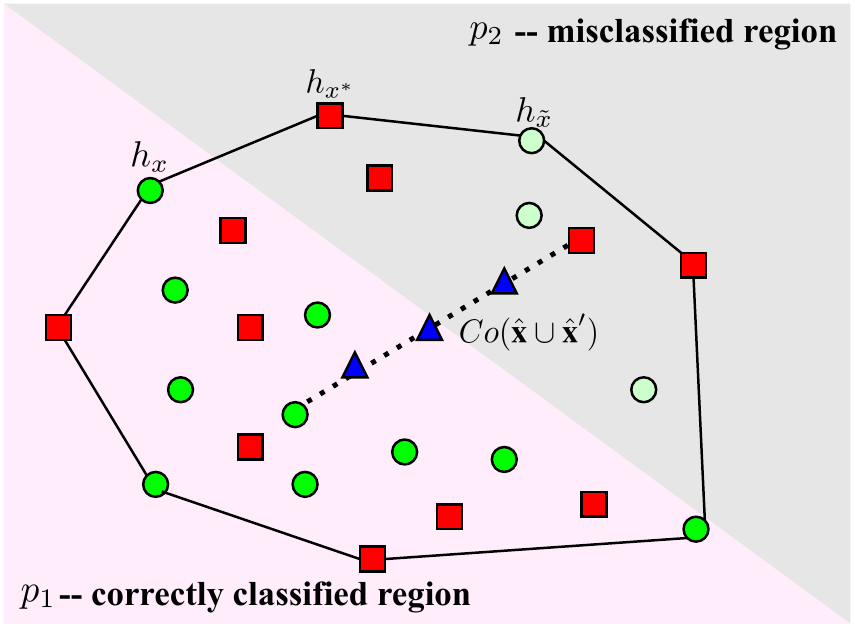}
         \caption{The schematic diagram of the data classification region and the convex hull.}
         \label{fig:convex}
    \end{minipage}%
    \hfill
    \begin{minipage}{0.45\textwidth}
        \centering
         \includegraphics[width=1\textwidth]{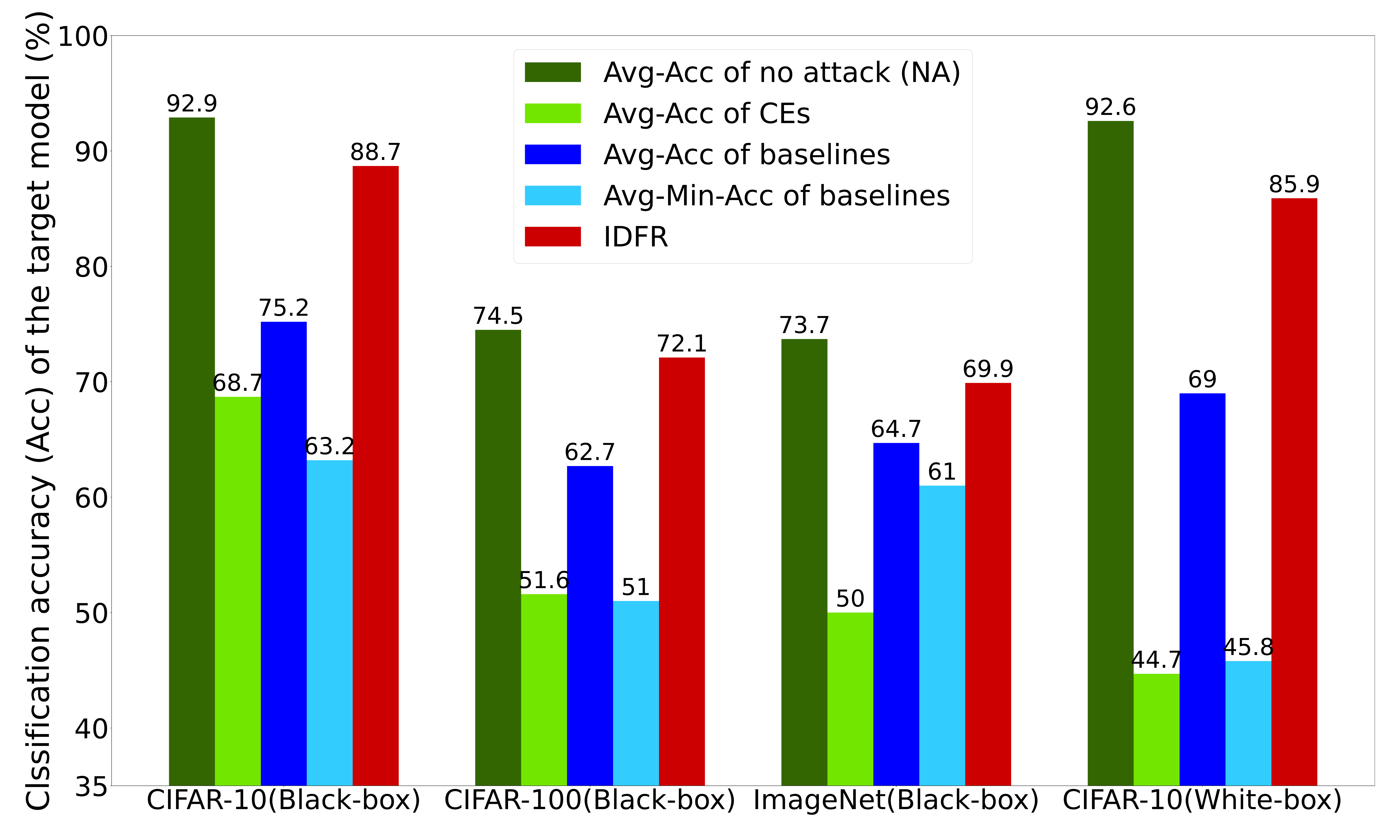}
         \caption{The statistical average performances  of the proposed IDFR and baselines.}
         \label{fig:res}
    \end{minipage}
\end{figure}

From the above experimental results shown in Table \ref{tab:black-cifar10} to Table \ref{tab:imagenet} and Figure \ref{fig:res}, we can draw the following conclusions: 1) Our proposed IDFR achieves a new SOTA (\underline{S}tate-\underline{O}f-\underline{T}he-\underline{A}rt) performance regardless of the attack target models and AEs. For example, our IDFR's average performance (classified accuracy of the target model) exceeds that of the baselines up to 13.3\%/9.4\%/5.8\% with the maximum average margin 25.5\%/21.1\%/8.9\% in different datasets and AEs. 2) Thanks to the novel low dimensional hidden lossy feature restoring of our FR, the performance of our IDFR is closest to that of the target model with clean samples as input, namely no defense, with only an average performance difference of 4.2\%/2.4\%/2.2\% in the various datasets and AEs, while the  difference for the baselines is as high as 17.7\%/11.8\%/9.0\% with the maximum average margin 25.5\%/21.1\%/11.9\%. Notably, Table \ref{tab:imagenet} clearly shows that our proposed IDFR reveals the optimal model adversarial robustness in the large-scale dataset ImageNet. 3) In general, the information loss of the defense attack model to the clean sample is huge, which leads to the sharp decrease of the target model performance (see Fig. \ref{fig:res}). This fact fully illustrates the correctness and importance of our observation:  an effective defense method against adversarial attacks must effectively deal with the key secondary disaster of the clean sample information loss after the denoising process.

\textbf{White-box attacks} In the experiments, the parameter settings of the various AEs generated by FGSM, CW and BIM are the same as in the white-box attack experiments. The  experimental results shown in Table \ref{tab:white-cifar10} clearly show that our IDFR is overwhelmingly superior to other baselines and can  effectively defend against a variety of white-box adversarial attacks.

\begin{table}[h]
    \centering
    \caption{The performance results on ImageNet with black-box attacks}
    \resizebox{0.98\textwidth}{12mm} {
        \begin{tabular}{lcccccccc}
         \toprule
          \multicolumn{1}{c}{Defense}  & \multicolumn{2}{c}{Clean Samples} & \multicolumn{2}{c}{FGSM($\epsilon=16$)} & \multicolumn{2}{c}{CW($\epsilon=16$)} &
          \multicolumn{2}{c}{BIM($\epsilon=16$)}\\ 
          \multicolumn{1}{c}{Method} & ResNet50 & InceptionV3 & ResNet50 & InceptionV3 & ResNet50 & InceptionV3 & ResNet50 & InceptionV3 \\ \midrule
          No Defense & 74.2 & 73.1 & 41.3 & 39.2 & 52.6 & 50.4 & 36.7 & 32.1 \\ 
          NRP & 66.7 & 63.8 & 62.1 & 63.8 & 66.4 & 68.4 & 59.3 & 59.0 \\ 
          Fast AT & 72.1 & 70.4 & 58.8 & 55.3 & 62.6 & 60.7 & 61.4 & 58.5 \\
          ComDefend & 68.3 & 66.4 & 68.3 & 67.9 & 70.7 & 69.1 & 65.3 & 63.1 \\ 
          IDFR(Ours) & \textbf{73.1} & \textbf{71.7} & \textbf{69.8} & \textbf{69.7} & \textbf{71.1} & \textbf{70.2} & \textbf{67.9} & \textbf{65.7} \\ \bottomrule
        \end{tabular}
    }
    \label{tab:imagenet}
\end{table}

\begin{table}[h]
    \centering
    \caption{The performance results on CIFAR-10 with white-box attacks}
    \resizebox{.84\textwidth}{22mm}{
    \begin{tabular}{lcccc}
    \toprule
      {Defense Method}  & Clean Samples & FGSM($\epsilon=4/8/16$) & CW($\epsilon=4/8/16$) &
      BIM($\epsilon=4/8/16$)\\ \midrule
      No Defense & 92.6 & 53.2/37.1/16.4 & 62.5/53.9/44.6 & 54.7/22.6/ 9.5 \\ 
      NRP & 77.6 & 75.8/73.6/69.6 & 76.8/74.1/70.4 & 78.2/77.9/76.8 \\ 
      Fast AT & 78.6 & 44.6/43.2/41.6 & 48.4/44.7/43.3 & 43.5/41.9/38.8 \\ 
      ComDefend & 85.7 & 80.2/77.7/72.3 & 82.9/81.3/80.7 & 81.1/77.3/73.9 \\ 
      Feature Squeezing & 91.3 & 81.8/74.2/57.7 & 83.7/81.9/75.8 & 80.7/59.1/48.4 \\ 
      JPEG(q=25) & 67.5 & 63.7/61.6/58.5 & 65.8/63.7/62.6 & 64.2/60.9/55.8 \\ 
      JPEG(q=50) & 74.6 & 69.3/62.7/57.9 & 71.2/69.6/67.4 & 68.3/62.7/50.9 \\ 
      JPEG(q=75) & 83.1 & 72.9/61.5/53.2 & 74.1/72.7/69.3 & 71.7/65.7/52.3 \\ 
      TVM & 87.5 & 76.9/63.1/53.6 & 73.9/71.8/64.2 & 77.5/73.8/65.2 \\ 
      ADP & 86.1 & 76.5/74.9/73.8 & 79.1/78.8/76.1 & 74.4/71.1/69.2 \\ 
      SOAP & 90.3 & 72.4/72.1/70.4 & 72.7/71.8/70.5 & 71.7/71.4/69.2 \\ 
      IDFR(Ours) & \textbf{92.1} & \textbf{86.8/84.4/78.1} & \textbf{88.2/86.8/86.5} & \textbf{87.6/85.6/82.8} \\ \bottomrule
    \end{tabular}
    }
    \label{tab:white-cifar10}
\end{table}



\textbf{Ablation Study} In this paper, we mainly adopt the jointly enhanced strategy of the input denoising (ID) and the hidden lossy feature restoring (FR) to improve the target model adversarial robustness. Table \ref{tab:alba} shows the experiments (black-box attacks running 20 iterations with $l_{\infty}$ bounded and  $\epsilon$=4/8/16 on target model ResNet50) of our method for ablation study, i.e., without FR, denoted as ID, or with jointly enhanced strategy both the ID and FR, denoted as ID+FR, respectively. The experiments clearly show that: 1) the ID model has strong ability of the adversarial denoising regardless of  datasets and AEs. For example, our ID improves the adversarial robustness of the target model by 34\% to 85\% in terms of various AEs, 2) the FR enhances the  adversarial robustness by 0.4\% to 16.8\% against various adversarial attacks, which fully verifies the necessity and importance of the FR, and 3) the jointly enhanced strategy of the ID+FR  achieves: (1) improved the target model adversarial robustness with a large margin, and (2) the increase tends of the model's defense performance with the increase of the adversarial attack powers, i.e., when $\epsilon $ increases. 


\vspace{-0.5em}
\begin{table}[h]
    \centering
    \caption{ The ablation experimental results on ResNet50 }
    \resizebox{.7\textwidth}{10mm}{
    \begin{tabular}{lcccc}
    \toprule
      \multirow{1}{*}{Defense}  &  \multicolumn{2}{c}{FGSM($\epsilon=4/8/16$)} & \multicolumn{2}{c}{CW($\epsilon=4/8/16$)} \\ 
      {Method} & CIFAR10 & SVHN & CIFAR10 & SVHN \\ \midrule
      No Defense & 52.7/44.0/34.0 & 52.1/40.0/27.4 &  17.1/14.2/ 9.0 & 37.1/17.2/ 9.5 \\
      ID & 86.7/90.5/92.2 & 92.6/95.3/96.1 & 81.8/82.1/73.2 & 95.8/93.5/94.6 \\ 
      ID+FR & \textbf{88.3/91.5/93.4} & \textbf{93.5/96.2/97.4} & \textbf{85.0/85.2/90.0} & \textbf{96.2/94.8/95.7} \\ \bottomrule
    \end{tabular}
    }
    \label{tab:alba}
\end{table}
\vspace{-1em} 
 \section{Conclusion}
In this study, to address the weakness of existing adversarial defense methods, for the first time, we introduced the formal and complete problem definition for the defense against adversarial attacks, and reveal the acoustic disaster, i.e., the decrease of the target model's  robustness due to the damaged clean samples caused by the AEs denoising. On the above basis, we presented a novel enhanced  adversarial defense method, namely IDFR, which consists of an enhanced input denoiser ID and an efficient  hidden lossy feature restorer FR with the convex hull optimization. Extensive experimental results  have verified the effectiveness of the proposed IDFR, and have clearly shown that the IDFR has  achieved  a new SOTA adversarial defense robustness performance compared to many state-of-the-art adversarial attack defense methods. How to further improve the defense performance of the proposed IDFR is our future research direction.
\newpage
\bibliography{iclr2022_conference}
\bibliographystyle{iclr2022_conference}

\newpage
\appendix
\section{Appendix}

\subsection{The network architectures of the proposed ID and FR  }\label{appendix:IDFRstructure}
In this paper, we introduce the U-Net convolutional network \citep{r19} as the basic network model of our proposed ID. The structure and  hyperparameters of the ID are shown in Table \ref{tab:ID-structure}. The ID   consists of 7 layers CNN, the outputs of the 3th layer and the 5th layer are connected to the input of the 6th layer, respectively. Similarly, the outputs of the 2th layer the 7th layer are connected to the input of the 8th layer, respectively. 

Similarly, as we discussed earlier, our FR model also uses the U-Net network model. The FR model consists of 6 layers fully connected networks as shown in Table \ref{tab:FR-structure}.

\begin{table}[h]
    \centering
    \caption{Hyperparameters of the ID Layers}
    \begin{tabular}{ccccc}
    \toprule
    layer & type & output channels & input channels & filter size \\ \midrule
    1st layer & conv+BN+ReLU & \ \ 32 & \ \ \ \ 3 & 3 $\times$ 3 \\
    2nd layer & conv+BN+ReLU & \ \ 32 & \ \ 32 & 3 $\times$ 3 \\
    3rd layer & conv+BN+ReLU & \ \ 64 & \ \ 32 & 3 $\times$ 3 \\
    4th layer & conv+BN+ReLU & \ \ 64 & \ \ 64 & 3 $\times$ 3 \\
    5th layer & conv+BN+ReLU & 128 & \ \ 64 & 3 $\times$ 3 \\
    6th layer & conv+BN+ReLU & 192 & 192 & 3 $\times$ 3 \\
    7th layer & conv+BN+ReLU & \ \ 64 & 192 & 3 $\times$ 3 \\
    8th layer & conv+BN+ReLU & \ \ 96 & \ \ 96 & 3 $\times$ 3 \\
    9th layer & conv+BN+ReLU & \ \  \ \ 3 & \ \ 96 & 3 $\times$ 3 \\
    \bottomrule
    \end{tabular}
    \label{tab:ID-structure}
\end{table}

\begin{table}[h]
    \centering
    \caption{Hyperparameters of the FR Layers}
    \begin{tabular}{ccccc}
    \toprule
    layer & type & output dimensions & input dimensions \\ \midrule
    1st layer & Fully Connect+ReLU & 256 & 512 \\
    2nd layer & Fully Connect+ReLU & 128 & 256 \\
    3rd layer & Fully Connect+ReLU & \ \ 32 & 128 \\
    4th layer & Fully Connect+ReLU & 128 & \ \ 32 \\
    5th layer & Fully Connect+ReLU & 256 & 128 \\
    6th layer & Fully Connect+ReLU & 512 & 256 \\
    \bottomrule
    \end{tabular}
    \label{tab:FR-structure}
\end{table}

\subsection{Proof of Theorem \ref{theorem1}}\label{appendix:proof}

\begin{proof} For the Theorem \ref{theorem1}, 
it is equal to the problem: given a convex hull $Co(\hat{\textbf{x}} \cup \hat{\textbf{x}}')$,  for any positive integer $n\geq2$, $\forall h_1, h_2, \cdots, h_n\in Co(\hat{\textbf{x}} \cup \hat{\textbf{x}}')$, given any non-negative real number $\alpha_1,\alpha_2,\cdots,\alpha_n$ and $\alpha_1+\alpha_2+\cdots+\alpha_n=1$, there is a constant $\alpha_1h_1+\alpha_2h_2+\cdots+\alpha_nh_n\in Co(\hat{\textbf{x}} \cup \hat{\textbf{x}}')$.

Next, we prove Theorem \ref{theorem1} by mathematical induction: 

By Eq. \ref{eq:convex-def}, the Theorem \ref{theorem1} holds when $n=2$.

Assume that the conclusion holds for $n=k$ points, and now we must prove that the conclusion holds for $n=k+1$ points.
\begin{equation}
    \because \forall \alpha_i \geq 0, \sum_{i=1}^{k+1}\alpha_i=1,\forall h_1, h_2, \cdots, h_n\in Co(\hat{\textbf{x}} \cup \hat{\textbf{x}}')
\end{equation}
Then, when $\sum_{i=1}^k\alpha_i = 0$ and  $\alpha_{k+1} = 1$, we have

\begin{equation}\label{eq:alpha-0}
h_{k+1}\in Co(\hat{\textbf{x}} \cup \hat{\textbf{x}}')
\end{equation}

$\because$ when $\sum_{i=1}^k\alpha_i \neq 0$,  
\begin{equation}\label{eq:alpha-neq-0}
\begin{aligned}
    \alpha_1h_1+ & \alpha_2h_2+\cdots+ \alpha_{k+1}h_{k+1}= \\
    & \Bigl( \sum_{i=1}^k \alpha_i \Bigl[ \frac{\alpha_1}{\sum_{i=1}^k\alpha_i}h_1+\frac{\alpha_2}{\sum_{i=1}^k\alpha_i}h_2+\cdots+\frac{\alpha_k}{\sum_{i=1}^k\alpha_i}h_k \Bigr] + \alpha_{k+1}h_{k+1} \Bigr)
\end{aligned}
\end{equation}

As the conclusion holds assuming for $n=k$ points, i.e., following Eq. \ref{eq:alpha-neq-0} holds
\begin{equation}\label{eq:proof-A}
     \Bigl( \sum_{i=1}^k \alpha_i \Bigl[ \frac{\alpha_1}{\sum_{i=1}^k\alpha_i}h_1+\frac{\alpha_2}{\sum_{i=1}^k\alpha_i}h_2+\cdots+\frac{\alpha_k}{\sum_{i=1}^k\alpha_i}h_k \Bigr] \Bigr) \in Co(\hat{\textbf{x}} \cup \hat{\textbf{x}}')
\end{equation}
$\because \alpha_{k+1}h_{k+1} \in Co(\hat{\textbf{x}} \cup \hat{\textbf{x}}')$,  $\sum_{i=1}^k\alpha_i\geq0, \alpha_{k+1}\geq0, \text{and} \sum_{i=1}^k\alpha_i+\alpha_{k+1}=1$.

Based on Eqs \ref{eq:alpha-0} and \ref{eq:proof-A}, with the basic definition of a convex set,  we can get the following Eq. \ref{eq:proof-final}.

\begin{equation}\label{eq:proof-final}
\alpha_1h_1+\alpha_2h_2+\cdots+\alpha_nh_n\in Co(\hat{\textbf{x}} \cup \hat{\textbf{x}}'),  n=k+1
\end{equation}
\end{proof}

\end{document}